# REVIEW. MACHINE LEARNING TECHNIQUES FOR TRAFFIC SIGN DETECTION

Rinat Mukhometzianov     Ying Wang

The Department of Electrical and Computer Engineering
The University of Waterloo

## ABSTRACT

An automatic road sign detection system localizes road signs from within images captured by an on-board camera of a vehicle, and support the driver to properly ride the vehicle. Most existing algorithms include a preprocessing step, feature extraction and detection step. This paper arranges the methods applied to road sign detection into two groups: general machine learning, neural networks. In this review, the issues related to automatic road sign detection are addressed, the popular existing methods developed to tackle the road sign detection problem are reviewed, and a comparison of the features of these methods is composed.

## 1. INTRODUCTION

World Health Organization reported that 1.25 million deaths happen each year due to traffic collisions. Static can climb up to 2 million each year with no appropriate precautionary actions are taken to make driving safer [1]. Distracted driving is the main reason for nearly 80% of traffic accidents [2]. We also should consider other factors including speeding, driver errors which frequently causes car crashes on roads [3].

One way to decrease number of driver's distractions is the usage of some Advanced Driver Assistance Systems (ADAS) [4]. One of the important component in modern ADAS is real-time computer vision processing units. Traffic sign detection and recognition algorithms is the crucial part of such systems [5].

The main aim of automated traffic signs detection and recognition methods is to localize and classify one or more road signs in a live video stream acquiring by a camera. The methods trying to implement on-board warning system to notify driver of upcoming meaningful traffic signs as quickly as possible to prevent possible car accident. These types of algorithms find applications in other areas, for example, in inventory management [6, 7], navigational map creation [8].

Creating a traffic sign detection and recognition system is a laborious vision task. Various important issues need to be resolved. These problems are featured below [9]:

- Visual obstacles due to season, weather conditions, etc.
- Moving vehicle usually the main reason for blurry images.
- It is impossible always have the same viewing angle of a road sign. Sign's pattern and shape generally distorted.
- Sign's colors may be faded because of weather conditions
- There could be multiple road signs on one frame
- Some signs can be deformed, (e.g. storm or pouring rain)
- Signs can vary in scale as the vehicles moving closer.
- Different artificial object of the same color, shape, size as traffic signs may be presented on a scene.
- The shape-based detection can be disturbed by some obstacles, such as billboard, pedestrian, trees, etc.
- Image quality varies from camera to camera.

The detection algorithm is the essential part of traffic sign recognition systems. Generally, detection algorithms are used in car assistance systems, that's why it is extremely important to achieve highest possible real-time speed of detection with highest achievable accuracy. All of those demonstrated that the field of road sign detection is challenging, crucial and alluring.

Paclik [10] mentioned that the earliest research on automated road sign detection and recognition started in 1984. A lot of research on detection and recognition was conducted since that year. A number of scientists have been working on traffic sign recognition and detection methods. For instance, Qian et al [11] achieved around 98% of detection rate.

Many working solutions have been suggested to enhance the performance of the automatic road sign detection and recognition system. Detection algorithm looks for traffic signs in an acquired image. Road signs represented in various shapes, colors, sizes and pattern. They formed to show its key elements through the consolidation of these three features.

Since 1989 [12] researchers started to apply machine learning methods for sign detection. Subject of machine learning is to study how to use computer imitate human learning activities, and to discover self-improvement methods of computers that to gather new skills and new knowledge, classify existing knowledge, and deliberately improve the performance [13].

Various method was proposed, such as, SVM, Viola-Jones, LDA, etc. Some methods showed significant performance. Area under ROC curve reaches 100% for some methods, but only for one database and no real-time [14]. Neural networks as a part of family of machine learning algorithms also used, e.g. Souza et al [15]. State-of-state approaches deep learning algorithms to achieve better performance, for example, Xiong et al [16].

Finally, we put all these machine learning methods in 3 main groups: general machine learning techniques, neural networks and deep learning networks.

## 2. REVIEW OF DATASETS

In traffic sign detection and recognition, it is very important to have suitable choices of datasets for different tasks and methods. Many different types of datasets have been collected and released to the public. Basically, these are some very commonly used datasets for different traffic sign detection and recognition systems, namely,

1) German Traffic Sign Detection Benchmark: 900 full images containing 1206 traffic signs. The images are selected from video sequences recorded near Bochum, Germany, capturing various environments such as urban, rural and highway during daylight and nighttime under diverse weather conditions. The images in the dataset were converted to RGB color space and stored in the format of raw PPM file. The road signs which are involved in the images were labelled manually [14].
2) Belgian Traffic Sign Dataset contains 9,000 still images with 13,444 road sign annotations matching with 4,565 different road signs which are visible within 50 meters from the camera. The images are produced by four video sequences captured by eight high resolution devices installed on the vehicle for recording several hours. The images are recorded at the urban areas from Flanders region in Belgium [17].
3) Chinese Traffic Sign Dataset: this dataset consists of 1100 images and the resolution of image are mainly 1024 * 768 and 1280 * 720. There are three categories of traffic signs inside the dataset, namely, prohibitory signs, danger signs and mandatory signs including challenging samples under various scenarios such as poor/strong lighting, affine transformation, imaging blurring, etc. [18]

In this review paper, there are also many other datasets of different regions and environment being used depending on different groups and research proposals, such as Google Street View panoramic images [8], Sweden Traffic Sign Detection Data Set [19], Tsinghua-Tencent 100K [20], TI Automotive Data Set [21], Road/Lane Detection Evaluation 2013 [22], 10 Video Clips of Streets Of Tokushima City [23].

## 3. MACHINE LEARNING

Machine learning techniques find their applications in many research areas [24]. Some of these methods widely used for detection and localization of objects. In case of road signs, they allow to find if image is a part one of them.

It is possible to divide all approaches in 2 classes. The first class presented by research that use methods like Viola-Jones, SVM, etc. directly to images. The second class is including methods which use different preprocessing techniques to extract features and then use them as an input data for machine learning methods.

One of the most popular class of algorithms is based on Support Vector Machine (SVM) which widely used in traffic sign detection systems. SVM are based on the concept of decision planes that separates between a set of objects having different class memberships and define decision boundaries.

Almost all methods use different preprocessing and feature extraction techniques before applying SVM. Timofte et al. [17] utilize Histogram of Oriented Gradients (HOG) and AdaBoost-selected Haar-like features before SVM with best False Negative (FN) 0.5%, False Positive (FP) 0.02% and 500 ms per 2 Megapixel frame that is hardly real-time performance. Moreover, they don't provide precision, recall or at least Area Under the Curve (AUC) for detection stage and characteristics of computer they made experiment on.

The other example of using HOG feature was proposed by Wang et al [25]. It is based on a hierarchical structure of SVM. The detection stage consists of pre-processing and SVM classifies input signs into different super classes like prohibitory signs, mandatory signs, etc. The performance with the accuracy of 99.89% and the average speed of 40 ms per image was achieved on GTSRB dataset, but not information about platform configuration and precision, recall estimations.

The authors in [26] also proposed a pipeline for traffic sign detection with HOG with adding MSER and WaDe algorithms for region extraction. The final performance is 98.15% of AUC and average processing time of ~840 ms per frame that is not acceptable for real-time, although information on the platform was not mentioned in the paper.

Another article which exploit HOG features along with Linear discriminant analysis (LDA) and SVM proposed by Wang et al [27]. They achieved precision and recall of 100% with GTSDB dataset and ~ 1.18 sec per frame on Core i3, 4 GB RAM MATLAB implemented algorithm which is not real-time speed at all.

As we can see MSER is also popular as HOG, for example we can see that Yang et al. [28] use them both with SVM and color probability model to achieve ~ 97.72% AUC, 0.162 sec per image speed on 3.7GHz CPU which show that this approach has a great potential for real-time application. They also provide recognition algorithm which show great accuracy. Unfortunately, they tested algorithm only with one dataset.

Last year (2016) research [29] proposed complex pipeline with segmentation, ROI candidate and PHOG/HOG feature extraction with binary SVM. They achieved 82% of accuracy, TP 87%, FP 23%, specificity 76%, precision 80%. Nonetheless they did not provide information about speed and platform.

Some recent paper [18] offers to use color gradients and shape detection as SVM input on Core i7 and 8 GB RAM with CTSD dataset. Sadly, no information about speed, accuracy, etc. were provided, so it impossible to say it this approach is really going to work.

Recent research like Shi and Lin [30] use bilateral Chinese transform (BCT), vertex, bisector transform (VBT) and HOG along with SVM to gain precision 88%, recall 90%. However, the paper has no information about speed and platform so it is hard to estimate if method is applicable to real time tasks. Training and testing dataset contain 5000 manually collected 512×288 pixels images without information of view angle, weather conditions, etc.

The second popular class of systems for traffic signs detection is based on AdaBoost. We have already reviewed one paper [17] which incorporate AdaBoost method. AdaBoost is a classification technique to construct strong classifier with linear combination of member classifiers, which improves the classification accuracy by increasing the weights of the misclassified data [31].

Chen and Lu [19] proposed complicated detection method which is consist of learning decision tree, AdaBoost, BoW histograms using the learned discriminative codewords and SVR model. They tested algorithm on 3 databases including GTSD, STSD, BTSD and got 99.92% of AUC at average, but no precision, recall evaluations are provided. The paper states that method process at average 0.05 to 0.5 sec for each frame on Core i7, 8GB RAM machine, so it gives opportunities to use it in real-time systems. Moreover, the algorithm can be used for classification, so for recognition.

Zang et al. [32] combined the Local Binary Pattern (LBP) feature detector with the AdaBoost Classifier implemented on OpenCV platform to extract regions of Interest for coarse selection, then used cascaded CNNs to reduce negative ROIs and make classification. They achieved the performance as on average 0.0074 sec for processing one sample image on Intel Core 2 Duo 2.2 GHz desktop computer with 4GB RAM, and 98.09% of AUC averagely.

One paper [21] offer to use AdaBoost with image pyramid and HOG features on low power embedded systems. No results for detection stage were provided, but they achieved 80% of accuracy, 33 ms per image on manually collected 1 MP dataset on 500 Mhz DSP for recognition stage which make it possible to run on real time systems and even integrate to existing platforms cause the methods run on DSP processors.

Researchers also proposed a lot of other road signs detection algorithms based on various machine learning techniques. For example, Kaplan et al. [33] use Adaptive Brightness Correction and genetic algorithms. With average 9 ms per frame of processing time on Intel T5450 computer they achieved 91% of true detections, 4.5% of false detections at average which was trained and tested on prerecorded 512x288 manually gathered videos. This performance makes this method potentially suitable for real-time, but precision and recall evaluations was not provided.

Some scientist suggested to use Bag of Visual Words with scale-invariant feature transform (SIFT) [34] which helps to achieve 89% of AUC in average on GTSDB dataset with 309 ms per image. Nevertheless, article is not including information about platform so it makes that hard to say if this method is suitable for real-time detection.

Huang et al. [35] proposed to use the color, the image segmentation and the hierarchical grouping methods based on selective search algorithm to detect the traffic sign. The selective search achieves the precision rate on different color space as 90.3% averagely, the correct rates for the detection system are 92.63%. No speed information is given, but claimed that it faster than search-based method

The positive side of most papers that they also introduce recognition techniques [17, 25, 28, 30, 31, 21, 33, 35, 36]. Moreover, some articles claim that speed could be significantly improved if rewrite algorithms with one of low-level programming language as C, C++, etc. [25, 27, 19]. The weakest point of nearly all papers that they evaluate algorithms only with one or two datasets. This create a lot of limitations on acceptable image resolution, quality and could not guarantee that all methods are universal in case of real world applications, for example for ADAS systems or possible mobile devices which could be a platform for sign detection software. Also, more the half of papers do not provide precision, recall or AUC evaluations which are important for such types of algorithms to find an optimal balanced in performance for practical applications [36].

## 4. ARTIFICIAL NEURAL NETWORK

Dr. Robert Hecht-Nielsen defines a neural network as: "*…a computing system made up of a number of simple, highly interconnected processing elements, which process information by their dynamic state response to external inputs*" [37]. Artificial Neural network is one group of algorithms which was biologically inspired and used for machine learning to model the data using graphs of Artificial Neurons, those neurons are a mathematical model that simulates approximately how a neuron in the brain works.

Neural networks showed outstanding performance in learning the input-output connections for nonlinear and complex task and systems [38]. It makes them a powerful and convenient tool for traffic sign detection.

Authors in [39] proposed traffic sign detection model based on Virtual Generalizing Random Access Memory Weightless Neural Networks (VG-RAM WNN) with very simple implementation and very fast training scheme. Running with an Intel Core i7, the model can be trained in 3.85 s using only 12 traffic signs, which is called a single TSD saccade. Detection speed: 3.6 sec per image. They mentioned that performance of the model could be improved with hardware accelerators, increasing training data, proper parameters tuning and so on.

The other approach is using of probabilistic neural networks (PNN) which was proposed by Zhang et al [40]. They combined adaptive color segmentation, central projection transformation (CPT) and PNN to achieve 95% success rate 2.8 s per image on common PC. Database is represented by $2452 \times 2056$ pixels of 510 images manually collected in Japan. Since no evaluation were provided on reshaped images this network we assume that this network is not suitable for real-time detection.

In the last 4 years, we can observe significant improvements in a field of so called Deep Learning algorithms. Over the recent few years, deep learning has had unprecedented success in fields like speech recognition, image classification, etc.

Deep learning is a subset of machine learning in Artificial Intelligence (AI) that has networks which are capable of learning unsupervised from data that is unstructured or unlabeled and is about learning multiple levels of representation and abstraction that help to make sense of data such as images, sound, and text [41]. So, it is make them very effective for traffic sign detection.

Wu et al. [42] proposed an approach where the image was firstly pre-processed to grey scale image by an SVM and then feed into some fixed and learnable layers of Convolutional Neural Networks. Running with the GTSDB dataset and Intel Xeon CPU E5620, 98.68% of precision-recall curve (AUC) of the testing dataset was achieved on average. Despite the accuracy, this model cannot do real-time detection since the processing time is quite long.

Authors in [20] provided schemes of two networks in total, one for detecting traffic sign alone, the other for simultaneously detection and recognition, both on the platform of Intel Xeon E5-1620, two NVIDIA Tesla K20 GPUs, 32GB RAM. Compared to the state-of-art Fast R-CNN with recall 0.56 and accuracy 0.50, their method achieved more significant result as 0.91 and 0.88 on recall and accuracy respectively. But the speed of this model is still not satisfied for real-time scenarios. They also proposed recognition system with comparatively good performance.

Xiong et al. [43] trained a traffic sign detection model based on deep CNNs using Region Proposal Network (RPN) in Fast R-CNN. Running on the hardware environment of NVIDIA GTX980Ti 6GB GPU, the average detection time is about 51.5 ms per image with the detection rate above 99% in continuous image sequence. The database they used is Chinese traffic sign with 7 main categories, so it is quite expected to see the performance on some other well-known dataset such as GTSDB. Even though no information about precision and recall was provided, it seems that this method is suitable for real-time detection. However, it might be tough to integrate it in ADAS system due to the high requirements of hardware.

New way to detect traffic signs was discovered by Peng et al. [44]. They use Faster R-CNN based on Region Proposal Networks to achieve 90% of accuracy with NVIDIA GTX 1070 8 GB GPU, Intel Core i5, 16 GB RAM on GTSDB dataset. The paper has no information about precision, recall. The method is required a lot of computational power, so it is may be hard to use it for real-time detection within, for instance, ADAS system.

Another Faster R-CNN-based model was proposed in [45], with two parts of using selective search to detect candidate regions firstly and then using CNNs to extract features, make classifications and modify parameters. They got the performance as 0.3449 on mAP value (Mean Average Precision) because some unsolved problems such as ignoring of signs when they are too small or the overlaps.

The most recent method suggested to use Image Pyramid with among bounding boxes. For all Fast R-CNN and Faster R-CNN the VGG-16 used for a feature representation. SOS-CNN [46] to achieve a recall of 0.93 and an accuracy of 0.90. However, no information about platform was provided.

## 5. CONCLUSION

Various commonly used machine learning techniques and neural network algorithms on road sign detection have been presented. Some of these techniques can be used with recognition algorithms to achieve real-time performance and even integrate them into ADAS.

In terms of datasets used in various kinds of traffic sign detection models, there are numbers of different choices. The most commonly used two datasets are GTSDB and GTSRB, since they are widely collected and well organized. There are also lots of algorithms using other datasets available such as BTSD, STSD [19] for training models or creating their own benchmark such as Tsinghua-Tencent 100K [20] for more complex real-world scenarios. Whether the model has feasibility in the real tasks or not somehow depends on the choice of datasets used for training process.

In machine learning algorithms, the most popular method is SVM. Since less papers on AdaBoost were presented it hard to directly compare it with other techniques. However, what we have is that comparing to second popular method (AdaBoost) the average speed of detection is lower while taking into consideration platforms algorithms were tested on, but SVM-based methods tends to be more stable and well researched. The best speed is achieved with combination of LBP and AdaBoost [32] with high AUC equal to 98.09%. It is tough task to directly compare performance of detection accuracy because only few papers have evaluation on several datasets. Moreover, all papers have missing part of evaluation features where it is precision, recall or AUC, or accuracy, TP, FP, etc. We can assume that method proposed by Wang et al. [27] have the gretest perfomance since precision and recall eqall to 100%.

Apart from various attempts related with machine learning methods, some researchers are inspired by the concept of neural network. The running speed of classical neural network algorithm seems not to be applicable in the real-time tasks. In last years, since significant progress has been proved in many other field with deep neural network, many researchers started to apply CNNs on traffic sign detection problem. More recently, R-CNNs combined with many other machine learning techniques have been attempted and some of the architectures gained magnificent improvements in accuracy. However, due to high expense in computation, deep neural network models have high hardware requirements, to achieve higher speed. Thus, it seems not easy to be used in ADAS systems.

Most results reported in the papers are tested on dataset of static images only, and only few on real road video recordings. Even though the traffic sign detection significantly improved in last 5 years, more research and tests on different datasets, real-time videos should be done. Research in traffic signs detection is significant and practical; it surely deserves more attention.